\documentclass{article}

\usepackage[final]{corl_2018}
\usepackage{amsthm}
\usepackage{graphicx}
\usepackage{bbm}
\usepackage{tikz,graphics,color,fullpage,float,epsf,caption,subcaption}
\usepackage{comment}
\usepackage{amssymb}
\usepackage{amsmath}
\usepackage{gensymb}

\makeatletter
\g@addto@macro\small{%
  \setlength\abovedisplayskip{-5pt}
  \setlength\abovedisplayshortskip{-5pt}
  \setlength\belowdisplayshortskip{-7pt}
  \setlength\belowdisplayskip{-7pt}
}
\makeatother

\usepackage{titlesec}
\titlespacing*{\section} {0pt}{0.7ex plus 0.2ex minus 0.2ex}{0.5ex plus .1ex minus 0.1ex}
\titlespacing*{\subsection} {0pt}{0.6ex plus 0.2ex minus 0.1ex}{0.4ex plus .1ex minus 0.1ex}
\titlespacing*{\paragraph} {0pt}{0.05ex plus 0.05ex minus .05ex}{1em}

 \usepackage[final]{corl_2018} %

\newcommand{\authorgap}{\hspace{2em}}
\author{
  Valts Blukis$^{\dagger\diamondsuit}$ \authorgap Dipendra Misra$^{\dagger\diamondsuit}$ \authorgap Ross A. Knepper$^{\dagger}$ \authorgap Yoav Artzi$^{\dagger\diamondsuit}$ \vspace{0.5em}\\
   $^{\dagger}$Department of Computer Science, Cornell University, Ithaca, New York, USA\\
$^{\diamondsuit}$Cornell Tech, Cornell University, New York, New York, USA\\
	\texttt{\{valts, dkm, rak, yoav\}@cs.cornell.edu}
}

\newtheoremstyle{mystyle}%
  {}%
  {}%
  {}%
  {}%
  {\itshape}%
  {.}%
  { }%
  {}%
\theoremstyle{mystyle}
\newtheorem{theorem}{Theorem}[section]

\newcommand{\eat}[1]{\ignorespaces}

\newcommand{\instrlen}{l}
\newcommand{\execlen}{T}
\newcommand{\idxtimestep}{t}

\newcommand{\trainsetsize}{N}
\newcommand{\testsetsize}{M}

\newcommand{\numdaggeriterations}{K}
\newcommand{\idxdaggeriteration}{k}

\newcommand{\allinstructions}{\mathcal{U}}

\newcommand{\state}{s}
\newcommand{\allstates}{\stateset}
\newcommand{\configuration}{\rho}
\newcommand{\position}{p}
\newcommand{\posseq}{\Xi}
\newcommand{\goalpos}{\position_g}

\newcommand{\trajvisit}{d^{\position}}
\newcommand{\stopvisit}{d^{g}}
\newcommand{\oracletrajvisit}{\trajvisit_*}
\newcommand{\oraclestopvisit}{\stopvisit_*}

\newcommand{\context}{c}

\newcommand{\act}[1]{{\tt #1}}
\newcommand{\action}{a}

\newcommand{\allactions}{\actionset}
\newcommand{\stopaction}{\act{STOP}}

\newcommand{\imagefunc}{\textsc{Img}}
\newcommand{\posefunc}{\textsc{Localize}}

\newcommand{\nlstring}[1]{{\em #1}}

\newcommand{\vanilladagger}{\textsc{DAgger}}
\newcommand{\daggerfm}{\textsc{DAggerFM}}
\newcommand{\policy}{\pi}
\newcommand{\oracle}{\pi^*}

\newcommand{\modelname}{\textsc{PVN}}

\newcommand{\distmodel}{\textsc{Visit}}
\newcommand{\actionmodel}{\textsc{Act}}
\newcommand{\stateset}{\mathcal{S}}

\newcommand{\transfunc}{\mathcal{T}}
\newcommand{\actionset}{\mathcal{A}}

\newcommand{\instruction}{u}
\newcommand{\instructionemb}{\mathbf{u}}

\newcommand{\params}{\theta}

\newcommand{\velfwd}{v}
\newcommand{\velang}{\omega}
\newcommand{\stopprob}{p^{{\rm stop}}}
\newcommand{\image}{I}
\newcommand{\pose}{P}
\newcommand{\rot}{\gamma}

\newcommand{\worldframe}{W}
\newcommand{\camframe}{C}

\newcommand{\fmcam}{\mathbf{F}^\camframe}

\newcommand{\fmworld}{\mathbf{F}^\worldframe}

\newcommand{\smworld}{\mathbf{S}^\worldframe}

\newcommand{\rmworld}{\mathbf{R}^\worldframe}

\newcommand{\smworldpos}{\tilde{\position}}

\newcommand{\weights}{\mathbf{W}}
\newcommand{\bias}{\mathbf{b}}

\newcommand{\resnet}{\textsc{ResNet} }

\newcommand{\lstm}{\textsc{LSTM} }

\newcommand{\lingunet}{\textsc{LingUNet}}

\newcommand{\leakyrelu}{\textsc{LeakyReLU}}

\newcommand{\gsmn}{\textsc{GSMN}}
\newcommand{\chaplot}{\textsc{Chaplot}}

\newcommand{\sysoracle}{\textsc{Oracle}}

\newcommand{\modelnoaux}{\textsc{PVN no aux}}
\newcommand{\modelnostop}{\textsc{PVN no $\stopvisit$}}
\newcommand{\modelprior}{\textsc{PVN no $\instruction$}}

\newcommand{\modelperfectact}{\textsc{PVN ideal $\actionmodel$}}
\newcommand{\modelperfectobs}{\textsc{PVN full obs}}

\newcommand{\oraclemodel}{\textsc{Oracle}}
\newcommand{\stopmodel}{\textsc{Stop}}
\newcommand{\avgmodel}{\textsc{Average}}

\newcommand{\policyoptimal}{\pi^*}

\newcommand{\navdrone}{\textsc{Lani}}

\newcommand{\numtrain}{19,758}
\newcommand{\numdev}{4,135}
\newcommand{\numtest}{4,072}
\newcommand{\numtotal}{27,965}

\newcommand{\avgnumsteps}{18}
\newcommand{\avgvel}{0.88}

\newcommand{\best}[1]{\textbf{#1}}

\newcommand{\lani}{\textsc{Lani}}

\newcommand{\featmap}{\mathbf{F}}
\newcommand{\featmaptxt}{\mathbf{G}}
\newcommand{\featmapdeconv}{\mathbf{H}}
\newcommand{\conv}{\textsc{CNN}}

\newcommand{\deconv}{\textsc{Deconv}}
\newcommand{\kernel}{\mathbf{K}}

\title{Mapping Navigation Instructions to Continuous Control Actions with Position-Visitation Prediction}

\begin{document}

\maketitle

\begin{abstract}
We propose an approach for mapping natural language instructions and raw observations to continuous control of a quadcopter drone. Our model predicts interpretable position-visitation distributions indicating where the agent should go during execution and where it should stop, and uses the predicted distributions to select the actions to execute. This two-step model decomposition allows for simple and efficient training using a combination of supervised learning and imitation learning. We evaluate our approach with a realistic drone  simulator, and demonstrate absolute task-completion accuracy improvements of 16.85\% over two state-of-the-art instruction-following methods.

\end{abstract}


\section{Introduction}
\label{sec:intro}

\begin{figure}
\centering
\begin{minipage}{.32\textwidth}
  \footnotesize
  \centering
  \frame{\includegraphics[width=\linewidth]{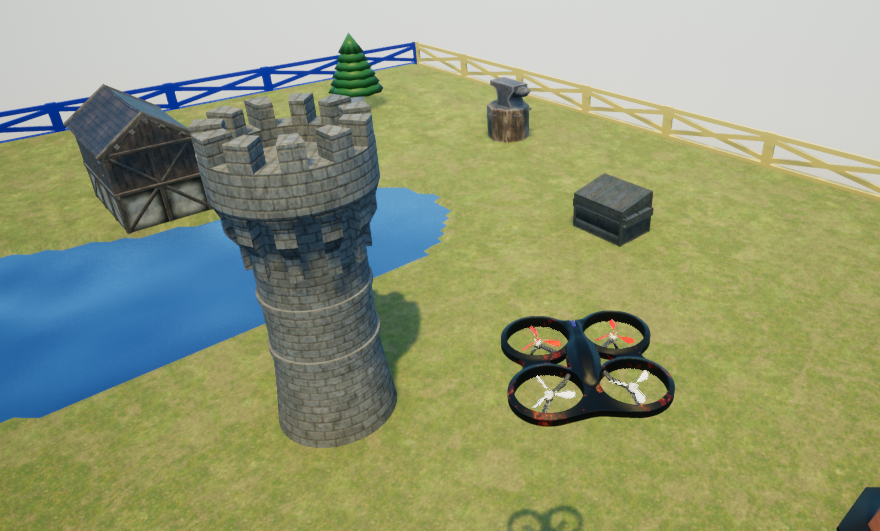}}
  \vskip 1pt \vskip 0pt
  \fbox{
  \begin{minipage}{0.935\linewidth}
  \textit{Go towards the blue fence passing the anvil and tree on the right}
  \end{minipage}}
  \caption{An example task from the $\navdrone$ dataset, showing the environment, agent, and the natural language instruction.}
  \label{fig:task}
\end{minipage}%
\hfill
\begin{minipage}{.65\textwidth}
  \centering
  \includegraphics[width=0.875\linewidth]{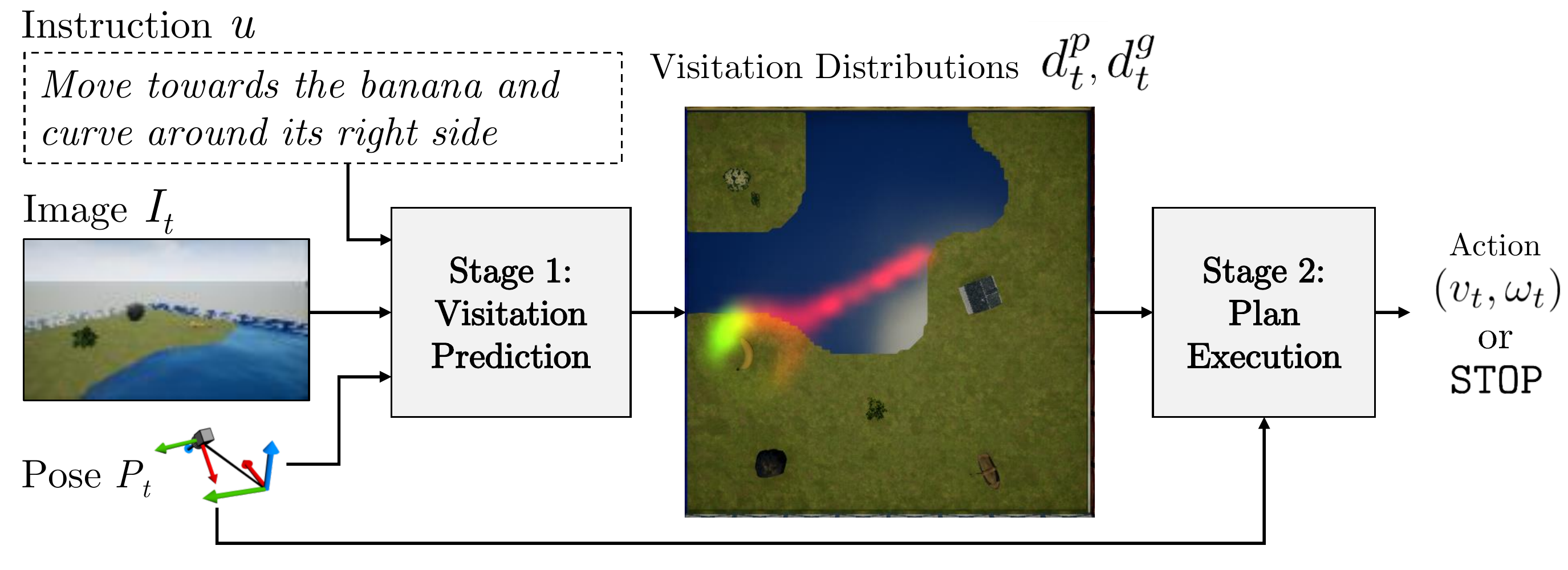}
  \caption{High-level illustration of our model. The model is decomposed into a visitation prediction component that predicts which areas in the environment to visit, and an execution component that outputs actions to drive the agent through predicted high-likelihood regions. The visitation prediction component predicts the positions to visit (red) and where to stop (green).}
  \label{fig:high_level}
\end{minipage}
\vspace{-10pt}
\end{figure}

Executing natural language navigation instructions from raw observations requires solving language, perception, planning, and control problems. 
Consider instructing a quadcopter drone using natural language. 
Figure~\ref{fig:task} shows an example instruction. 
Resolving the instruction requires identifying the \nlstring{blue fence}, \nlstring{anvil} and \nlstring{tree} in the world, understanding the spatial constraints \nlstring{towards} and \nlstring{on the right}, planning  a trajectory that satisfies these constraints, and continuously controlling the quadcopter to follow the trajectory. 
Existing work has addressed this problem mostly using manually-designed symbolic representations for language meaning and environment~\cite{huang2010natural, tellex11grounding, matuszek2012learning, thomason2015learning, arumugam2017accurately, gopalan2018sequence}. This approach  requires significant knowledge representation effort and is hard to scale. 
Recently, \citet{blukis2018following} proposed to trade-off the representation design with representation  learning. 
However, their approach was developed using synthetic language only, where a small set of words were combined in a handful of different ways. 
This does not convey the full complexity of natural language, and may lead to design decisions and performance estimates  divorced from the real problem.

In this paper, we study the problem of executing instructions with a realistic quadcopter simulator using a corpus of crowdsourced natural language navigation instructions. 
Our data and environment  combine language and robotic challenges. 
The instruction language is rich with linguistic phenomena, including object references, co-references within sentences, and spatial and temporal relations; the environment simulator provides a close approximation of realistic quadcopter flight, including a realistic  controller that requires rapid decisions in response to  continuously changing observations.

We address the complete execution problem with a single model that is decomposed into two stages of planning and plan execution. Figure~\ref{fig:high_level} illustrates the two stages. 
The first stage takes as input the language and the  observations of the agent, and outputs two distributions that aim to to solve different challenges: (a) identifying the positions that are likely to be visited during a correct instruction execution, and (b) recognizing the correct goal position. 
The second stage of the model controls the drone to fly between the high probability positions to complete the task and reach the most likely goal location.
The two stages  are combined into a single neural network. 
While the approach does not require designing an intermediate symbolic representation, the agent plan is still interpretable by simple visualization of the distributions over a map.

Our approach introduces two learning challenges: (a) estimate the model parameters with the limited language data available and a realistic number of experiences in the environment; and (b) ensure the different parts of the model specialize to solve their intended tasks. 
We address both challenges by training each of the two stages separately. 
We train the  visitation prediction stage with supervised learning using expert demonstrations, and  the plan execution stage by mapping expert visitation distributions to actions using imitation learning. 
At test time,  the second stage uses the predicted distributions from the first. 
This learning method emphasizes sample efficiency. 
The first stage uses supervised learning with training demonstrations; the second stage is independent from the complex language and visual reasoning required, allowing for sample-efficient imitation learning. 
This approach also does not suffer from the credit assignment problem of training the complete network using rewards on actions only. This ensures the different parts of the network solve their intended task, and the generated interpretable distributions are representative of the agent reasoning.

To evaluate our approach, we adapt the $\lani$ corpus~\cite{misra2018mapping} for the realistic quadcopter simulator from~\citet{blukis2018following}, and create a continuous control instruction following benchmark. The $\lani$ corpus includes \numtotal{} crowdsourced natural language instructions paired with human demonstrations. 
We compare our approach to the continuous-action analogs of two recently proposed approaches~\cite{misra2017mapping,chaplot2017gated}, and demonstrate absolute task-completion accuracy improvements of 16.85\%. 
We also discuss a generalization of our position visitation prediction approach to state-visitation distribution prediction for sequential decision processes, and suggest the conditions for applying it to future work on robot learning problems.  
The models, dataset, and environment are publicly available at \url{https://github.com/clic-lab/drif}.

\begin{figure}
\label{fig:first}
\end{figure}

\section{Technical Overview}
\label{sec:tech}

\paragraph{Task}
Let $\allinstructions$ be the set of natural language instructions, $\allstates$ be the set of world states, and $\allactions$ be the set of all actions. An instruction $\instruction$ is a sequence of $\instrlen$ tokens $\langle \instruction_1, \dots, \instruction_\instrlen \rangle$. 
An action $\action$ is either a tuple $(\velfwd, \velang)$ of forward and angular velocities or the completion action $\stopaction$. The state $\state$ contains information about the current configuration of all objects in the world. 
Given a start state $\state_1 \in \allstates$ and an instruction $\instruction \in \allinstructions$, the agent executes $\instruction$ by generating a sequence of actions, where the last action is the special action $\stopaction$, which indicates task completion.
The agent behavior  is determined by its configuration $\configuration$.
An execution of length $\execlen$ is a sequence $\langle (\state_1, \action_1), \dots, (\state_{\execlen}, \action_{\execlen}) \rangle$, where $\state_\idxtimestep \in \allstates$ is the state at timestep $\idxtimestep$, $\action_{\idxtimestep} \in \allactions$ is the action updating the agent configuration, and the last action is $\action_{\execlen} = \stopaction$.
Given an action $\action_{\idxtimestep} = (\velfwd_{\idxtimestep}, \velang_{\idxtimestep})$, we set the agent configuration $\configuration = (\velfwd_{\idxtimestep}, \velang_{\idxtimestep})$, which specifies the controller setpoint.  
Between actions, the agent maintains its configuration.

\paragraph{Model}
The agent does not have access to the world state. 
At timestep $\idxtimestep$, the agent observes the agent context $\context_{\idxtimestep} = (\instruction, \image_1,\cdots, \image_{\idxtimestep}, \pose_1, \cdots \pose_{\idxtimestep})$, where $\instruction$ is the instruction and  $\image_i = \imagefunc (\state_i)$ and $\pose_i = \posefunc (\state_i)$, $i=1 \dots \idxtimestep$ are  monocular first-person RGB images and 6-DOF agent poses observed at time step $i$. 
The pose $\pose_{i}$ is a pair $(\position_{i}, \rot_{i})$, where $\position_{i}$ is a position and $\rot_{i}$ is an orientation. 
Given an agent context $\context_{\idxtimestep}$, we predict two visitation distributions that define a plan to execute and the actions required to execute the plan. 
A visitation distribution is a discrete distribution over positions in the environment. 
The trajectory-visitation distribution $\trajvisit$ puts high probability on positions in the environment the agent is likely to go through during execution, and the goal-visitation distribution $\stopvisit$ puts high probability on positions where the agent should $\stopaction$ to complete its execution.  
Given $\trajvisit$ and $\stopvisit$, the second stage of the model predicts the actions to complete the task by going through high probability positions according to the distributions. 
As the agent observes more of the environment during execution, the distributions are continuously updated.

\paragraph{Learning}
We assume access to a training set of $\trainsetsize$ examples $\{ (\instruction^{(i)}, \state_1^{(i)},  \posseq^{(i)})\}_{i = 1}^{\trainsetsize}$, where $\instruction^{(i)}$ is an instruction, $\state_1^{(i)}$ is a start state, and $\posseq^{(i)} = \langle \position_1^{(i)}, \dots, \position_{\execlen}^{(i)}\rangle$ is a sequence of  positions that defines a trajectory generated from a human demonstration execution of $\instruction$. 
Learning is decomposed into two stages. We first train the visitation distributions prediction  given the visitation distributions inferred from the oracle policy $\oracle$. We then use imitation learning using $\oracle$ to generate the sequence of actions required given the visitation distributions.

\paragraph{Evaluation}

We evaluate  on a test set of $\testsetsize$ examples $\{ (\instruction^{(i)}, \state_1^{(i)},  \goalpos^{(i)}) \}_{i = 1}^{\testsetsize}$, where $\instruction^{(i)}$ is an instruction, $\state_1^{(i)}$ is a start state, and $\goalpos^{(i)}$ is the goal position. 
We consider the task successfully completed if the agent outputs the $\stopaction$ action within a predefined Euclidean distance of  $\goalpos^{(i)}$. We additionally evaluate the mean and median Euclidean distance to the goal position. 

\section{Related Work}
\label{sec:related}

Natural language instruction following has been studied extensively on physical robots~\cite{matuszek2012joint, duvallet2013imitation, walter2013learning, tellex11grounding, misra2014context, duvallet2013imitation, hemachandra2015learning, knepper15} and simulated agents~\cite{macmahon2006walk, branavan2010reading, matuszek2010following, matuszek2012learning,artzi2013weakly,Suhr:18situated}. 
These methods require hand-engineering of intermediate symbolic representations, an effort that is hard to scale to complex domains.  
Our approach does not require a symbolic representation, instead relying on a learned spatial representation induced directly from human demonstrations. 
Our approach is related to recent work on executing instructions without such symbolic representations using discrete environments~\cite{misra2017mapping,shah2018follownet,anderson2017vision,misra2018mapping}. 
In contrast, we use  a continuous environment. 
While we focus on the challenge of using natural language, this problem was also studied using synthetic language with the goal of abstracting natural language challenges and focusing on navigation~\cite{hermann2017grounded, chaplot2017gated} and continuous control~\cite{blukis2018following}. 
Our approach is related to recent work on learning visuomotor control policies for grasping~\cite{lenz2015deep, levine2016learning, quillen2018deep}, dexterous manipulation~\cite{levine2016end, nair2017combining, tobin2017domain} and visual navigation~\cite{bhatti2016playing}. 
While these methods have mostly focused on learning single robotic tasks, or transferring a single task between multiple domains~\cite{tobin2017domain, srinivas2018universal, bousmalis2017using, tan2018sim}, our aim is to train a model that can execute navigation tasks specified using natural language, including previously unseen tasks during test time. 

Treating planning as prediction of visitation probabilities is related to recent work on  neural network models that explicitly construct internal maps~\cite{gupta2017cognitive, blukis2018following, khan2018memory}, incorporate external maps~\cite{bhatti2016playing, savinov2018semi}, or do planning~\cite{srinivas2018universal}. 
These architectures take advantage of domain knowledge to provide sample-efficient training and interpretable representations. 
In contrast, we cast planning as an image-to-image mapping~\cite{ronneberger2015u, zhu2017unpaired}, where the output image is interpreted as a probability distribution over environment locations.
Our architecture borrows building blocks from prior work. 
We use the ResNet architecture for perception~\cite{he2016deep} and the neural mapping approach of \citet{blukis2018following} to construct a dynamic semantic map. 
We also use the $\lingunet$ conditional image translation module~\cite{misra2018mapping}. 
While it  was introduced for first-person goal location prediction, we use it to predict  visitation distributions. 

Learning from Demonstrations (LfD) approaches have previously decomposed robot learning into learning high-level tasks and low-level skills (e.g. Dynamic Movement Primitives~\cite{pastor2009learning, pastor2011skill, konidaris2012skilltrees, maeda2017corl}). Our approach follows this general idea. 
However, instead of using trajectories or probabilities as task representations~\cite{parasachos2013nips}, we predict visitation distributions using a neural network. 
This results in a reactive approach that defers planning of the full trajectory and starts task execution under uncertainty that is gradually reduced with additional observations. 
This approach does not assume access to the full system configuration space or a symbolic environment representation.
Furthermore, the learned representation is not constrained to a specific robot. For example, the same predicted visitation distribution could potentially be used on a humanoid or a ground vehicle, each running its own plan execution component.

\section{Model}
\label{sec:model}

\begin{figure*}[t]
\centering
\includegraphics[scale=0.24]{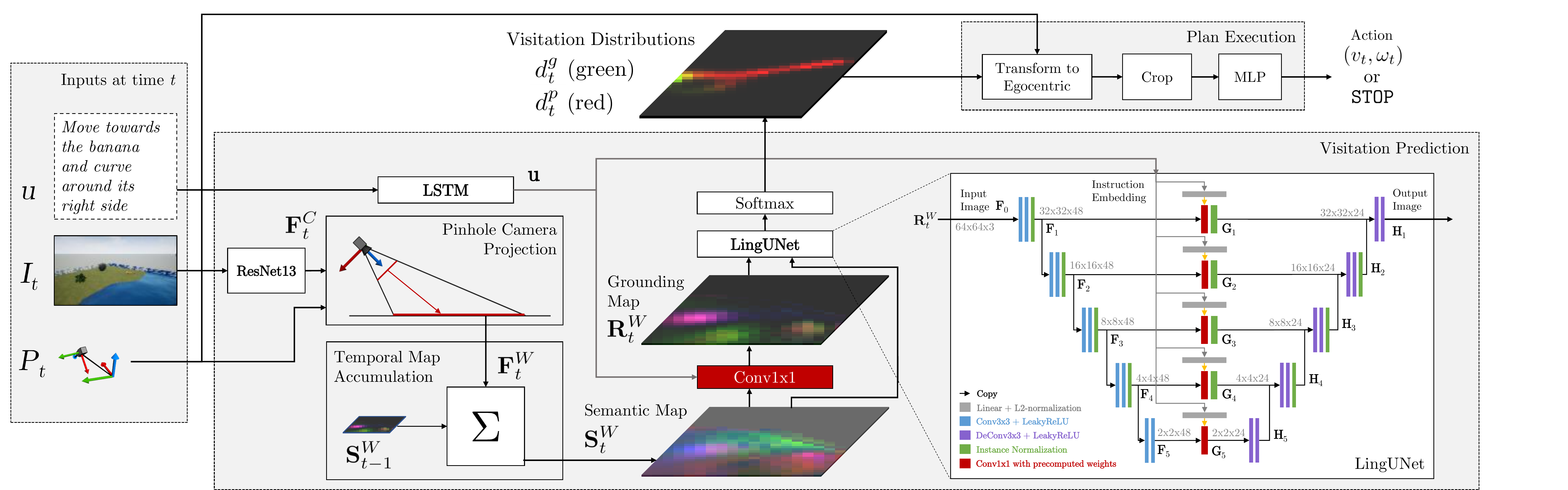}
\caption{An illustration of our model architecture. The instruction $\instruction$ is encoded into an instruction embedding $\instructionemb$ using an $\lstm$ network. At each timestep $t$, image features $\fmcam_t$  are produced with a custom residual network~\cite[ResNet;][]{he2016deep}, projected to the world reference frame through a pinhole camera model, and accumulated through time into a globally persistent semantic map $\smworld_t$. The map is  used to predict the visitation distributions $\trajvisit$ and $\stopvisit$ by using $\instructionemb$ to create a grounding map $\rmworld_t$ and generate the distributions using the $\lingunet$ architecture. A simple execution network then transforms the distributions to an egocentric reference frame and generates the next action.}
\label{fig:arch}
\end{figure*}

\begin{figure*}[t]
\centering
\includegraphics[width=0.89\linewidth]{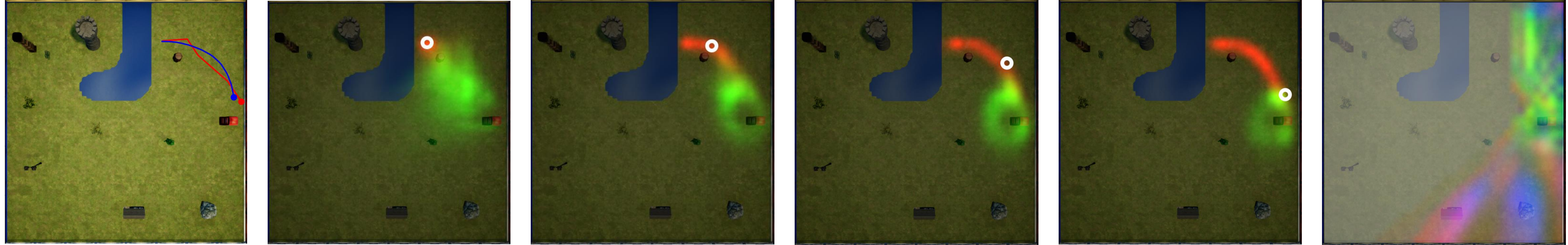}
\caption{Predicted visitation distributions for the the instruction from the $\navdrone$ development set \nlstring{go around the barrel and then move towards the phone booth}. The left-most image shows top-down view with the human demonstration trajectory (red) and our model trajectory (blue). The next four images are predicted visitation distributions, $\trajvisit$ (red) and $\stopvisit$ (green), as the execution progresses (left to right). The white circle represents agent's current position. %
The uncertainty over the stopping position decreases with time as the semantic map accumulates more information. The right image shows the first three channels of the final semantic map $\smworld_{\execlen}$.}
\label{fig:dist_evolution}
\end{figure*}

We model the agent behavior using a neural network policy $\policy$. 
The input to the policy at time $\idxtimestep$ is the agent context $\context_{\idxtimestep} = (\instruction, \image_1,\cdots, \image_{\idxtimestep}, \pose_1, \cdots \pose_{\idxtimestep})$, where $\instruction$ is the instruction and  $\image_i = \imagefunc (\state_i)$ and $\pose_i = \posefunc (\state_i)$, $i=1 \dots 
\idxtimestep$ are  first-person images and 6-DOF agent poses observed at timestep $i$ and state $\state_i$. 
The policy outputs an action $\action_\idxtimestep = (\velfwd_{\idxtimestep}, \velang_{\idxtimestep})$, where $\velfwd_\idxtimestep$ is a forward velocity and $\velang_\idxtimestep$ is an angular velocity, and a probability for the $\stopaction$ action $\stopprob_\idxtimestep$. 
We decompose the policy to visitation prediction and plan execution. 
Visitation prediction $\distmodel(\context_\idxtimestep)$ computes a 2D discrete semantic map $\smworld_t$. Each position in $\smworld_t$ corresponds to an area in the environment, and represents it with a learned vector.  
The map is used to generate two probability distributions: trajectory-visitation distribution  $\trajvisit_\idxtimestep(\smworldpos \mid \context_\idxtimestep )$ and goal-visitation distribution $\stopvisit_\idxtimestep(\smworldpos \mid \context_\idxtimestep)$, where $\smworldpos$ is a position in $\smworld_t$. 
The first distribution models the probability of visiting each position as part of an optimal policy executing the instruction $\instruction$, and the second the probability of each position being the goal where the agent should select the $\stopaction$ action. 
We update the semantic map $\smworld_\idxtimestep$ at every timestep with the latest  observations. The distributions $\trajvisit_\idxtimestep$ and $\stopvisit_\idxtimestep$ are only computed every $T_d$ timesteps. 
When not updating the distributions, we set $\trajvisit_\idxtimestep = \trajvisit_{\idxtimestep-1}$ and $\stopvisit_\idxtimestep = \stopvisit_{\idxtimestep - 1}$. This allows for periodic re-planning and limits the computational workload.
In the second stage,  plan execution $\actionmodel(\trajvisit_\idxtimestep, \stopvisit_\idxtimestep, \pose_\idxtimestep )$ generates the action $\action_t$ and the stop  probability $\stopprob_t$. 
Figure~\ref{fig:arch} illustrates our architecture, and Figure~\ref{fig:dist_evolution} shows  example visitation distributions generated by our approach.

\subsection{Stage 1: Visitation Prediction}

\paragraph{Feature Projection and Semantic Mapping}

We predict the visitation distributions over a learned semantic map of the environment.
We construct the map using the method of \citet{blukis2018following}. 
The full details of the process are specified in the original paper. 
Roughly speaking, the semantic mapping process includes three steps: feature extraction, projection, and accumulation.
At timestep $\idxtimestep$, we process the currently observed image $\image_\idxtimestep$ using a 13-layer residual neural network $\resnet$ to generate a feature map $\fmcam_\idxtimestep = \resnet(\image_\idxtimestep)$ of size $W_f \times H_f \times C$.
We compute a feature map in the world coordinate frame $\fmworld_\idxtimestep$ by projecting $\fmcam_\idxtimestep$ with a pinhole camera model onto the ground plane at elevation zero. 
The semantic map of the environment $\smworld_\idxtimestep$ at time $t$ is an integration of $\fmworld_\idxtimestep$ and $\smworld_{\idxtimestep-1}$, the  map from the previous timestep. 
The integration equation is given in Section 4c in \citet{blukis2018following}. 
This process generates a tensor  $\smworld_{\idxtimestep}$  of size $W_w \times H_w \times C$ that represents a map, where each location $[\smworld_{\idxtimestep}]_{(x,y)}$ is a $C$-dimensional feature vector computed from all past observations $\image_{<\idxtimestep}$, each processed to learned features $\fmcam_{<\idxtimestep}$ and  projected onto the environment ground in the world frame at coordinates $(x,y)$. 
This map maintains a learned high-level representation for every world location $(x,y)$ that has been visible in any of the previously observed images.
We define the world coordinate frame using the agent starting pose $\pose_1$: the agent position is the coordinates $(0,0)$, and the positive direction of the $x$-axis is along the agent heading. 
This gives consistent meaning to spatial language, such as \nlstring{turn left} or \nlstring{pass on the left side of}.

\paragraph{Instruction Embedding}
We represent the instruction $\instruction = \langle \instruction_1, \cdots \instruction_{\instrlen} \rangle$ as an embedded vector $\instructionemb$. We generate a series of hidden states $\mathbf{h}_i = \textsc{LSTM}(\phi(\instruction_i), \mathbf{h}_{i-1})$, $i = 1\dots \instrlen$, where $\textsc{LSTM}$ is a Long-Short Term Memory~\cite{hochreiter1997long} Recurrent Neural Network (RNN) and $\phi$ is a learned word-embedding function. The instruction embedding is  the last hidden state $\instructionemb = \mathbf{h}_{\instrlen}$.

\paragraph{Position Visitation Distribution Prediction} 

We use image generation to predict the visitation distributions $\trajvisit_\idxtimestep$ and $\stopvisit_\idxtimestep$. For each of the two distributions, we generate a matrix of dimension $W_w \times H_w$, the height and width dimensions of the semantic map $\smworld_\idxtimestep$, and normalize the values to compute the distribution. 
To generate these matrices we use $\lingunet$, a language-conditioned image-to-image encoder-decoder architecture~\cite{misra2018mapping}.

The input to $\lingunet$ is the semantic map $\smworld_\idxtimestep$ and a grounding map $\rmworld_t$ that incorporates the instruction $\instruction$ into the semantic map.
We create $\rmworld_t$ with a 1$\times$1 convolution $\rmworld_\idxtimestep = \smworld_\idxtimestep \circledast \kernel_G$. The kernel $\kernel_G$ is computed using a learned linear transformation  $\kernel_G = \weights_G \instructionemb + \bias_G$, where $\instructionemb$ is the instruction embedding. 

The grounding map $\rmworld_\idxtimestep$ has the same height and width as $\smworld_\idxtimestep$, and during training we optimize the parameters so it captures the objects mentioned in the instruction $\instruction$ (Section~\ref{sec:learn}). 

$\lingunet$ uses a series of convolution and deconvolution operations.
The input map $\featmap_0 = [\smworld_\idxtimestep, \rmworld_\idxtimestep]$ is processed through $L$ cascaded convolutional layers to generate a sequence of feature maps $\featmap_k = \conv_k(\featmap_{k-1})$, $k = 1\dots L$.\footnote{$[\cdot,\cdot]$ denotes concatenation along the channel dimension.}
Each  $\featmap_k$ is filtered with a 1$\times$1 convolution with weights $\kernel_k$. The kernels $\kernel_k$ are computed from the instruction embedding $\instructionemb$ using a learned linear transformation $\kernel_k = \weights^{u}_{k} \instructionemb + \bias^{u}_{k}$. 
This generates $l$ language-conditioned feature maps $\featmaptxt_k = \featmap_k \circledast \kernel_k$, $k = 1\dots L$.
A series of $L$ deconvolution operations computes $L$ feature maps of increasing size:

\begin{small}
\vspace{-6pt}
\begin{eqnarray*}
\featmapdeconv_k = \left\{\begin{array}{lr}
        {\deconv_k([\featmapdeconv_{k+1}, \featmaptxt_{k}])}, & \text{if } 1 \leq k \leq L-1\\
        {\deconv_k(\featmaptxt_{k})}, & \text{if } k=L
        \end{array}\right. \;\;,   
\end{eqnarray*}
\end{small}

The output of $\lingunet$ is $\featmapdeconv_1$, which is of size $W_w \times H_w \times 2$. 
The full details of $\lingunet$ are specified in \citet{misra2018mapping}.
We apply a softmax operation on each channel of $\featmapdeconv_1$ separately to generate the trajectory-visitation distribution $\trajvisit_t$ and the goal-visitation distribution $\stopvisit_t$. 
In Section~\ref{sec:learn}, we describe how we estimate the parameters to ensure that $\trajvisit$ and $\stopvisit$ model the visitation distributions.

\subsection{Stage 2: Plan Execution}

The action generation component $\actionmodel(\trajvisit_\idxtimestep, \stopvisit_\idxtimestep, \pose_\idxtimestep)$ generates the action values from the two visitation distributions $\trajvisit_\idxtimestep$ and $\stopvisit_\idxtimestep$ and the current agent pose $\pose_\idxtimestep$.  
We first perform an affine transformation of the most recent visitation distributions to align them with the current agent egocentric reference frame as defined by its pose $\pose_\idxtimestep$, and crop a $K\times K$ region centered around the agent's position. 
We fill the positions  outside the semantic map with zeros. 
We flatten and concatenate  the cropped regions of the distributions into a single vector $\mathbf{x}$ of size $2K^2$, and compute the feed-forward network: 

\vspace{-3pt}
\begin{small}
\begin{eqnarray*}
e^{\rm stop}_\idxtimestep, \velfwd_\idxtimestep, \velang_\idxtimestep &=& \weights^{(2)}[\mathbf{x}; \leakyrelu\{0, \weights^{(1)}\mathbf{x} + \bias^{(1)}\}]  + \bias^{(2)} \;\;,
\end{eqnarray*}
\end{small}
\vspace{-3pt}

\noindent
where $[;]$ denotes concatenation of vectors and \leakyrelu{} is a leaky ReLU non-linearity~\cite{maas2013rectifier}.
If the stopping probability $\stopprob_\idxtimestep = \sigma(e^{\rm stop}_\idxtimestep)$ is above a threshold $\kappa$ the agent takes the stop action. Otherwise, we set the controller configuration using the forward velocity $\velfwd_\idxtimestep$ and angular velocity $\velang_\idxtimestep$.

\section{Learning}
\label{sec:learn}

Our model parameters can be divided into two groups. 
The visitation prediction $\distmodel(\cdot)$ parameters $\params_1$ include the parameters of the functions $\phi$, $\lstm$, and $\resnet$, $\weights_G$, $\bias_G$, and the components of $\lingunet$: $\{\conv_k\}_{k=1}^{L}$, $\{\deconv_k\}_{k=1}^{L}$, $\{\weights_k^u\}_{k=1}^{L}$, $\{\bias_k^u\}_{k=1}^{L}$.  
The plan execution $\actionmodel(\cdot)$ parameters $\params_2$ are $\weights^{(1)}$, $\bias^{(1)}$, $\weights^{(2)}$, $\bias^{(2)}$. 
We use supervised learning to estimate the visitation prediction parameters and imitation learning for the plan execution parameters.

\paragraph{Estimating  Visitation Prediction Parameters} 

We assume access to training examples $(\instruction, \state_1,  \posseq)$, where $\instruction$ is an instruction, $\state_1$ is a start state, and $\posseq= \langle \position_1, \dots, \position_{\execlen}\rangle$ is a sequence of $\execlen$ positions.\footnote{To simplify notation, we describe learning for a single example.}
We convert the sequence $\posseq$ to a sequence of positions in the semantic map $\tilde{\posseq} = \langle \smworldpos_1, \dots, \smworldpos_{\execlen}\rangle$. 
We generate expert trajectory-visitation distribution $\oracletrajvisit$ by assigning high probability for positions around the demonstration trajectory, and goal-visitation distribution $\oraclestopvisit$  by assigning high probability around the goal position $\position_T$. 
For each location $\smworldpos = (x,y)$ in the semantic map, we calculate the probability of visiting and stopping there as:

\begin{small}
\begin{equation*}
\oracletrajvisit(\smworldpos) = \frac{1}{Z_p} \sum_{\smworldpos_t \in \tilde{\posseq}} g(\smworldpos | \smworldpos_t, \sigma)
\quad \quad
\oraclestopvisit(\smworldpos) = \frac{1}{Z_g} g(\smworldpos | \smworldpos_T, \sigma)  \;\;,
\end{equation*}
\end{small}

\noindent
where $g(\cdot | \mu, \sigma)$ is a Gaussian probability density function with mean $\mu$ and variance $\sigma^2$, and $Z_p$ and $Z_g$ are normalization terms. 
The distributions are computed efficiently by applying a Gaussian filter on an image of the human trajectory. 
We then generate a sequence of agent contexts $\context_t$ by executing an oracle policy $\policyoptimal$, which is implemented with a simple control rule that steers the quadcopter along the human demonstration trajectory $\posseq$. 
We create a training example $( \context_t,  \oracletrajvisit, \oraclestopvisit)$ for each time step $t = 1,T_d+1, 2T_d+1, \dots$ in the oracle trajectory when we compute the visitation distributions, and minimize the KL divergence between the expert and predicted distribution: \mbox{$D_{KL}(\oracletrajvisit \mid\mid \trajvisit( \cdot \mid \context_t)) + D_{KL}(\oraclestopvisit \mid\mid \stopvisit( \cdot \mid \context_t))$}. 
The data and objective do not consider the incremental update of the distributions, and we always optimize towards the full visitation distributions. 

We additionally use three auxiliary loss functions from \citet{blukis2018following}  to bias the different components in the model to specialize as intended: 
(a) the object recognition loss  $J_{\rm percept}$  to classify visible objects using their corresponding positions in the semantic map;
(b) the grounding loss $J_{\rm ground}$ to classify if a visible object in the semantic map is mentioned in the instruction $\instruction$; 
and (c) the language loss $J_{\rm lang}$ to classify if objects are mentioned in the instruction $\instruction$. 
To compute $J_{\rm  ground}$ and $J_{\rm lang}$, we use alignments between words and object labels that we heuristically extract from the training data using pointwise mutual information. Please refer to the supplementary material for full details.

The complete objective for an example $( \context_t,  \oracletrajvisit, \oraclestopvisit)$ for time $t$ is:

\begin{small}
\begin{multline}
\nonumber J(\params_1) = D_{KL}(\oracletrajvisit \mid\mid \trajvisit( \cdot \mid \context_t)) + D_{KL}(\oraclestopvisit \mid\mid \stopvisit( \cdot \mid \context_t)) \: + \\
+ \lambda_{\rm percept} J_{\rm percept}(\params_1) + \lambda_{\rm ground} J_{\rm ground}(\params_1) + \lambda_{\rm lang} J_{\rm lang}(\params_1)\;\;,
\label{eqn:stage1}
\end{multline}
\end{small}

where $\lambda_{(\cdot)}$ is a hyperparameter weighting the contribution of the corresponding auxiliary loss.

\paragraph{Estimating Plan Execution Parameters} 

We train the plan execution stage $\actionmodel(\trajvisit, \stopvisit, \pose)$ using imitation learning with the oracle policy $\policyoptimal$. 
During imitation learning, we use the visitation distributions $\oracletrajvisit$ and $\oraclestopvisit$ induced from the human demonstrations. 
This provides the model access to the same information that guides the oracle policy, which it learns to imitate. 
We use $\daggerfm$~\cite{blukis2018following}, a variant of $\vanilladagger$~\cite{ross2011reduction} for low-memory usage. 
$\daggerfm$ performs $\numdaggeriterations$ iterations of training. 
For each iteration $\idxdaggeriteration$ and a training example $(\instruction, \state_1, \posseq)$, we generate an execution $\langle (\state_1, \action_1),  (\state_2, \action_2) \cdots (\state_T, \action_T) \rangle$ using a mixture policy. 
The mixture policy selects an action at time $t$ using  $\policyoptimal$ with probability $\beta^\idxdaggeriteration$ or the learned policy $\actionmodel(\oracletrajvisit, \oraclestopvisit, \pose_t)$ with probability $1-\beta^\idxdaggeriteration$, where $\beta \in (0, 1)$ is a hyperparameter. 
The states generated in the execution are aggregated in a dataset across iterations. After each iteration, we prune the dataset to a fixed size and perform one epoch of supervised learning. We use a binary cross-entropy loss for the $\stopaction$ probability $\stopprob$, and a mean-squared-error loss for the velocities. When the oracle selects $\stopaction$, both velocities are zero. 
We initialize imitation learning with supervised learning using the oracle policy $\policyoptimal$ trajectories.

\paragraph{Discussion} 

Our approach is an instance of learning state-visitation distributions in Markov Decisions Processes (MDP). 
Consider an MDP $\langle \stateset, \actionset, R, \transfunc, H, \mu \rangle$, where $\stateset$ is a set of states, $\actionset$ is a set of actions, $R: \stateset  \rightarrow [0, R_{max}]$ is a reward function, $\transfunc: \stateset \times \actionset \rightarrow Pr(\stateset)$ is a probabilistic transition function, $H$ is the time horizon, and $\mu$ is the start-state distribution.\footnote{$Pr(\cdot)$ denotes a probability distribution.}
The state-visitation distribution of a policy $\pi: \stateset \rightarrow Pr(\actionset)$ is defined as $d(s; \pi, \mu) = \frac{1}{H}\sum_t d_t(\state; \policy, \mu)$, where $d_t(\state; \policy, \mu)$ is the probability of visiting state $\state$ at time $t$ following policy $\policy$ with the initial state-distribution $\mu$. 

Reasoning about the entire state space $\stateset$ is  challenging. 
Instead, we consider an alternative discrete state space $\tilde{\stateset}$ with a mapping $\phi: \stateset \rightarrow \tilde{\stateset}$  and a reward function $\tilde{R}: \tilde{\stateset} \rightarrow \mathbbm{R}^+$. 
For example, in a robot navigation scenario,  $\tilde{\state}$ can be  the robot pose estimate $\tilde{\state} = \pose$, or the positions in our semantic map $\smworld$. In a manipulation setup, $\tilde{\state}$ can be  the manipulator configuration. This choice is task-specific, but should include variables that are are measurable and relevant to task completion. 
The state-visitation distribution in $\tilde{\stateset}$ is $\tilde{d}(\tilde{s}; \pi, \mu) = \int \mathbbm{1}\{\phi(s) = \tilde{s}\} d(s; \pi, \mu) ds$. 
In general, we construct $\tilde{S}$ as a small set to support efficient computation of the visitation distribution, and enable our two stage learning. In the first stage, we train a visitation model to predict the visitation distribution $\tilde{d}(\cdot; \pi^*, \mu)$ for the oracle policy $\pi^*$,  and in the second stage, we learn a plan execution model using the oracle visitation distribution $\tilde{d}(\cdot; \pi^*, \mu)$ using imitation learning.%

There is a strong relation between learning the state distribution and policy learning. 
For predicted visitation distributions $\hat{d}$ with a bounded error in regard to the optimal visitation distribution, the sub-optimality error of policies that accurately follow the predicted distribution is bounded as well:
\begin{theorem}
Suppose $D_{KL}(\tilde{d}(\cdot; \pi^*,\mu) \mid \mid \hat{d}) \le \epsilon$ and let $\Pi(\eta) = \{\pi \mid D_{KL}(\hat{d} \mid \mid \tilde{d}(\cdot; \pi, \mu)) \le \eta\}$ be the set of all policies whose approximate state-visitation distribution has at maximum $\eta$ KL divergence from $\hat{d}$. Assume that for every $s \in \stateset$ there holds $|R(s) - \tilde{R}(\phi(s))| \le \alpha$. 
Then: 

\begin{small}
\begin{equation}
\sup_{s \in \stateset} \sup_{\pi \in \Pi(\eta)} V^*(s) - V^\pi(s) \le H\left(R_{max} + \alpha\right)\left(\sqrt{2 \epsilon} + \sqrt{2\eta}\right)\;\;.  \nonumber
\end{equation}
\end{small}
\end{theorem}

\section{Experimental Setup}
\label{sec:evaluation}

\paragraph{Data and Environments}
We evaluate our approach on the $\navdrone$ corpus~\cite{misra2018mapping}. $\navdrone$ contains $\numtotal$ crowd-sourced instructions for navigation in an open environment. 
Each datapoint includes an instruction, a human-annotated ground-truth demonstration trajectory, and an environment with various landmarks and lakes. The dataset train/dev/test split is \numtrain/\numdev/\numtest.
Each environment specification defines placement of 6--13 landmarks within a square grass field of size 50m$\times$50m.
We use the quadcopter simulator environment from \citet{blukis2018following} based on the Unreal Engine,\footnote{https://www.unrealengine.com/} which uses the AirSim plugin~\cite{airsim2017fsr} to simulate realistic quadcopter dynamics. %

\paragraph{Data Augmentation}
We create additional data for visitation prediction learning by rotating the semantic map $\smworld$ and the gold distributions, $\oracletrajvisit$ and $\oraclestopvisit$,  by a random angle $\alpha \sim \mathcal{N}(0, 0.5\text{rad})$. This allows the agent to generalize beyond the common behavior of heading towards the object in front.

\paragraph{Evaluation Metric}
We measure the stopping distance of the agent from the goal as \mbox{$\Vert \goalpos - \position_\execlen \Vert$}, where $\goalpos$ is the end-point of the human annotated demonstration and $\position_\execlen$ is the position where the agent output the $\stopaction$ action. 
A task is completed successfully if the stopping distance is \mbox{$<5.0m$},  10\% of the environment edge-length. 
We also report the average and median stopping distance. 

\paragraph{Systems}
We compare our Position-visitation Network ($\modelname$) approach to the $\chaplot$~\cite{chaplot2017gated} and $\gsmn$~\cite{blukis2018following} approaches. $\chaplot$ is an instruction following model that makes use of gated attention. 
Similar to our approach, $\gsmn$ builds a semantic map, but uses simple language-derived convolutional filters to infer the goal location instead of computing visitation probabilities. 
We also report $\sysoracle$ performance as an upper bound and two trivial baselines: (a) $\textsc{Stop}$: stop immediately;  and (b) $\textsc{Average}$: fly forward for the average number of steps ($\avgnumsteps$) with the average velocity ($\avgvel m/s$), both computed with the $\sysoracle$ policy from the training data.
Hyperparameter settings are provided in the supplementary material.

\begin{table}[t]
\begin{center}
\begin{minipage}{.48\linewidth}
\vspace{-2pt}
\footnotesize
\centering
\begin{tabular}{|l|c|c|c|c|}
\hline
Method & SR (\%) & AD & MD  \\
\hline
\hline
\multicolumn{4}{|l|}{\textbf{Test Results}} \\
\hline
\hline
\stopmodel         & \phantom{00}5.72        & 15.8\phantom{0}        & 14.8\phantom{0}          \\
\avgmodel          & \phantom{0}16.43       & 12.5\phantom{0}		   & 10.1\phantom{0}          \\
\hline
\chaplot           & \phantom{0}21.34       & 11.2\phantom{0}        & \phantom{0}9.35          \\
\gsmn          	   & \phantom{0}24.36       & \phantom{0}9.94        & \phantom{0}8.28          \\
\best{\modelname}  & \phantom{0}\best{41.21} & \phantom{0}\best{8.68} & \phantom{0}\best{6.26}   \\
\hline
\oraclemodel       & 100.0\phantom{0}       & \phantom{0}1.38        & \phantom{0}1.29          \\
\hline
\hline
\multicolumn{4}{|l|}{\textbf{Development Ablations and Analysis}} \\
\hline
\hline
\best{\modelname} & \best{40.44} & \phantom{0}\best{8.56}  & \phantom{0}\best{6.28}    \\
\hline
\modelnoaux          & 30.77     & 10.1\phantom{0}        & \phantom{0}7.94          \\
\modelnostop         & 35.98     & \phantom{0}9.25        & \phantom{0}7.2\phantom{0}          \\
$\modelname \textsc{ no } \vanilladagger$ & 38.87 & \phantom{0}9.18 & \phantom{0}6.69  \\ 
\modelprior          & 23.07     & 11.6\phantom{0}        & 10.1\phantom{0}          \\
\hline
\modelperfectact	 & 45.70	 & \phantom{0}8.42		   & \phantom{0}6.25 		   \\
\modelperfectobs	 & 60.59	 & \phantom{0}5.67 		   & \phantom{0}4.0\phantom{0}	   \\
\hline
$\modelname$ $\text{height} \div 2$	 & 39.51	 & \phantom{0}8.95		   & \phantom{0}6.55 			   \\
$\modelname$ $\velang \times 2$	 & 41.09	 & \phantom{0}8.6\phantom{0} 		   & \phantom{0}6.12			   \\
\hline
\end{tabular}
\caption{Test and development results, including model analysis. We evaluate success rate (SR), average stopping distance (AD), and median stopping distance (MD).}
\label{tab:test_results}
\end{minipage}
\hfill
\begin{minipage}{.48\linewidth}

\begin{center}
\includegraphics[width=\linewidth]{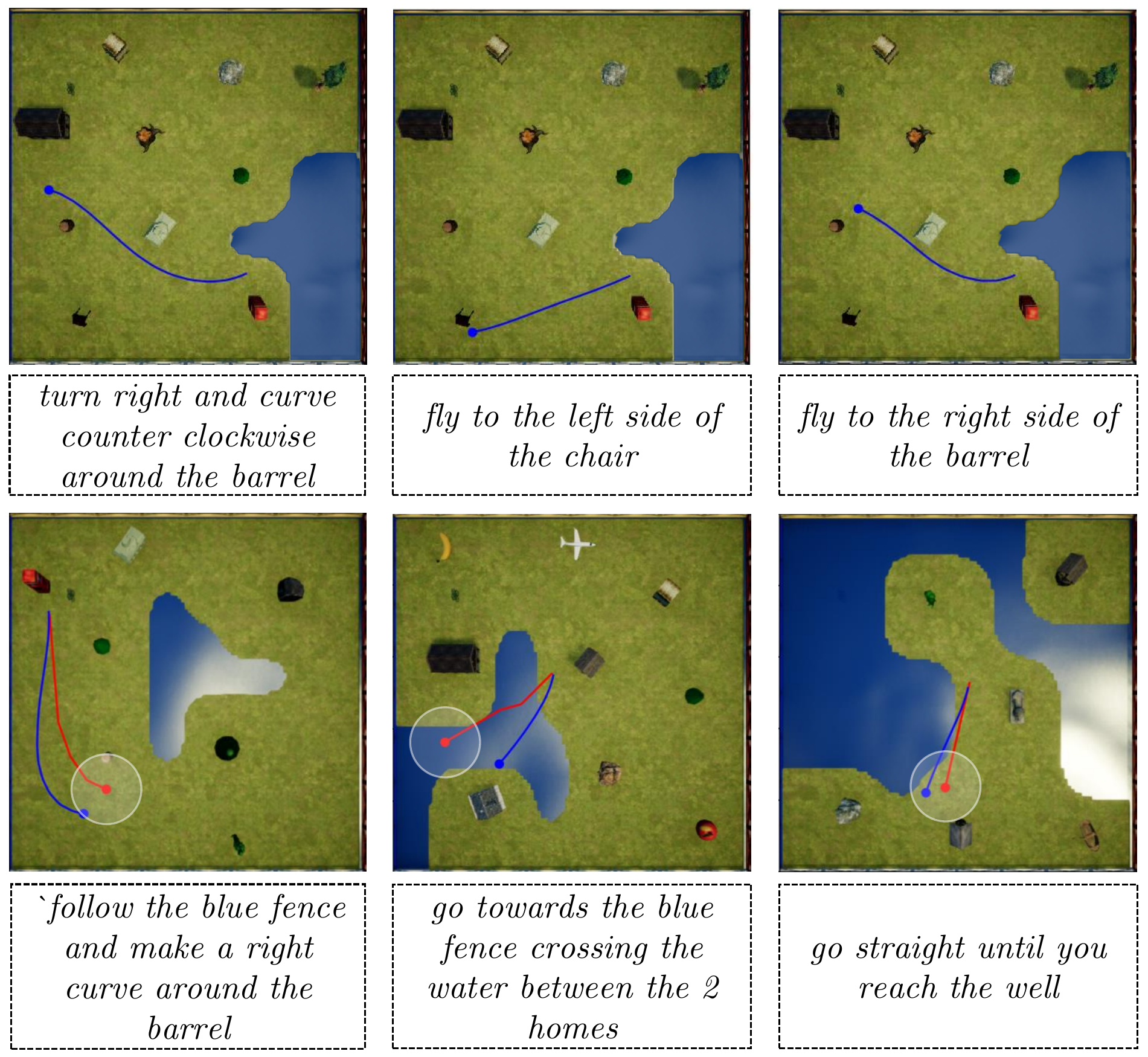}
\captionof{figure}{Our model executing engineered instructions from a single starting position (top) and representative instructions from the $\navdrone$ development set (bottom). The maps show human demonstrations (red), our model trajectories (blue), and goal regions (white circles).}
\label{fig:instructions}
\end{center}
\end{minipage}
\end{center}
\vspace{-30pt}
\end{table}

\section{Results}
\label{sec:results}

Table~\ref{tab:test_results} shows the performance on the test set and our ablations on the development set. 
The  low performance of the $\stopmodel$ and $\avgmodel$ baselines shows the hardness of the problem. 
Our full model $\modelname$ demonstrates absolute task-completion improvement of 16.85\%  over the second-best system ($\gsmn$), and a relative improvement of 12.7\% on average stopping distance and 32.3\% on the median stopping distance. 
The relatively low performance of $\gsmn$ compared to previous results with the same environment but synthetic language~\cite{blukis2018following}, an accuracy drop of 54.8, illustrates the challenges introduced by natural language. 
The performance of $\chaplot$ similarly degrades by 9.6 accuracy points compared to previously reported results on the same corpus but with a discrete environment~\cite{misra2018mapping}. This demonstrates the challenges introduced by a realistic simulation. 

Our ablations show that all components of the methods contribute to its performance. 
Removing the auxiliary objectives ($\modelnoaux$) or the goal-distribution prediction to rely only on the trajectory-visitation distribution ($\modelnostop$) both lower performance significantly. 
While using imitation learning shows a significant benefit, model performance degradation  is less pronounced when only using supervised learning for the second stage ($\modelname \textsc{ no } \vanilladagger$). 
The low performance of the model without access to the instruction ($\modelprior$) illustrates that our model makes effective use of the input language. 
Figure \ref{fig:instructions} shows example trajectories executed by our model, illustrating the ability to reason  about spatial language. 
The supplementary material includes more examples.

We evaluate the quality of goal-visitation distribution $\stopvisit$ with an ideal plan execution model that stops perfectly at the most likely predicted stopping position $\smworldpos_g = \arg \max \stopvisit(\smworldpos)$. 
The performance increase from using  a perfect goal-visitation distribution  with our model ($\modelperfectact$) illustrates the improvement that could be achieved by a better plan execution policy.
We observe a more drastic improvement with full observability ($\modelperfectobs$), where  the input image $\image_\idxtimestep$ is set to the top-down view of the environment.
This suggests  the model architecture is capable of significantly higher performance with improved exploration and mapping.

Finally, we do initial tests for model robustness against test-time variations. 
We test for visual differences by flying at 2.5m ($\modelname$ $\text{height} \div 2$), half the training height (5.0m). 
We test for dynamic differences by doubling the angular velocity during testing for every output action ($\modelname$ $\velang \times 2$). 
In both cases, the difference in model performance is relatively small, revealing the robustness of a modular approach to small visual and dynamics differences.

\section{Conclusion}
\label{sec:conclusion}

We study the problem of mapping natural language instructions and raw observations to continuous control of a quadcopter drone. 
Our approach is tailored for navigation. We design a model that enables interpretable visualization of the agent plans, and a learning method optimized for sample efficiency. 
Our modular approach is suitable  for related tasks with different robotics agents. 
However, the effectiveness of our mapping mechanism with limited visibility, for example with a ground robot, remains to be tested empirically in future work. 
Investigating the generalization of our visitation prediction approach to other tasks also remains an important direction for future work.  
 

\section*{Acknoledgements}

This research was supported by Schmidt Sciences, NSF award CAREER-1750499, AFOSR award FA9550-17-1-0109, the Amazon Research Awards program, and cloud computing credits from Amazon.
We thank the anonymous reviewers for their helpful comments.

{\footnotesize
\setlength{\bibsep}{4pt}
\bibliography{references}
}

\clearpage
\appendix
\section{Details on Auxiliary Objectives}

We use three additive auxiliary objectives to help the different components of the model specialize as intended with limited amount of training data. 

\paragraph{Object Recognition Loss}

The object-recognition objective $J_{\rm percept}$ ensures  the semantic map $\smworld_\idxtimestep$ stores information about locations and identities of various objects.
At timestep $\idxtimestep$, for every object $o$ that is visible in the first person image $\image_\idxtimestep$, we classify the element in the semantic map $\smworld_\idxtimestep$ corresponding to the object location in the world.
We apply a linear softmax classifier to every semantic map element that spatially corresponds to the center of an object. At a given timestep $\idxtimestep$ the classifier loss is: 

\begin{small}
\begin{equation*}
J_{\rm percept}(\params_1) = \frac{-1}{|O_{{\rm FPV}}|}\sum_{o \in O_{{\rm FPV}}}[\hat{y}_o log(y_o)]\;\;,
\end{equation*}
\end{small}

where $\hat{y}_o$ is the true class label of the object $o$ and $y_o$ is the predicted probability. $O_{{\rm FPV}}$ is the set of objects visible in the image $\image_\idxtimestep$.

\paragraph{Grounding Loss}

For every object $o$ visible in the first-person image $\image_\idxtimestep$, we use the feature vector from the grounding map $\rmworld_\idxtimestep$ corresponding to the object location in the world with a linear softmax classifier to predict whether the object was mentioned in the instruction $u$. The objective is:

\begin{small}
\begin{equation*}
J_{\rm ground}(\params_1) = \frac{-1}{|O_{{\rm FPV}}|}\sum_{o \in O_{{\rm FPV}}}[\hat{y}_o log(y_o) + (1-\hat{y}_o) log(1 - y_o)]\;\;,
\end{equation*}
\end{small}

where $\hat{y}_o$ is a 0/1-valued label indicating whether the object o was mentioned in the instruction and $y_o$ is the corresponding model prediction. $O_{{\rm FPV}}$ is the set of objects visible in the image $\image_\idxtimestep$.

\paragraph{Language Loss}

The instruction-mention auxiliary objective uses a similar classifier to the grounding loss. Given  the instruction embedding $\instructionemb$, we predict for each of the 63 possible objects whether it was mentioned in the instruction $u$. The objective is:

\begin{small}
\begin{equation}
\nonumber J_{\rm lang}(\params_1) = \frac{-1}{|O|}\sum_{o \in O_{{\rm FPV}}}[\hat{y}_o log(y_o) + (1-\hat{y}_o) log(1 - y_o)]\;\;,
\end{equation}
\end{small}

where $\hat{y}_o$ is a 0/1-valued label, same as above.

\section{Automatic Word-object Alignment Extraction}

In order to infer whether an object $o$ was mentioned in the instruction $u$, we use automatically extracted word-object alignments from the dataset. Let $E(o)$ be the event that an object $o$ occurs within 15 meters of the human-demonstration trajectory ${\posseq}$, let $E(\tau)$ be the event that a word type $\tau$ occurs in the instruction $\instruction$, and let $E(o, \tau)$ be the event that both $E(o)$ and $E(\tau)$ occur simultaneously. The pointwise mutual information between events $E(o)$ and $E(\tau)$ over the training set is:

\begin{small}
\begin{equation*}
  \textsc{PMI}(o, \tau) = P(E(o,\tau)) \log \frac{P(E(o,\tau))}{P(E(o))P(E(\tau))}\;\;,
\end{equation*}
\end{small}

where the probabilities are estimated from counts over training examples $\{ (\instruction^{(i)}, \state_1^{(i)},  \posseq^{(i)})\}_{i = 1}^N$.
The output set of word-object alignments is:

\begin{small}
\begin{equation*}
  \{(o, \tau) \: | \: \textsc{PMI}(o, \tau) > T_{\rm PMI} \wedge P(\tau) < T_{\tau}\}\;\;,
\end{equation*}  
\end{small}

where $T_{PMI} = 0.008$ and $T_{\tau} = 0.1$ are threshold hyperparameters.

\section{Hyperparameter Settings}

\paragraph{Image and Feature Dimensions}\mbox{}\\
Camera horizontal FOV: $90\degree$\\
Input image dimensions: $128 \times 72 \times3$\\
Feature map $\fmcam$ dimensions: $32 \times 18 \times 32$\\
Semantic map $\smworld$ dimensions: $64 \times 64 \times 32$\\
Visitation distributions $\stopvisit$ and $\trajvisit$ dimensions: $64 \times 64 \times 1$\\
Cropped visitation distribution dimensions: $12 \times 12 \times 1$\\
Environment edge length in meters: $50m$\\
Environment edge length in pixels on $\smworld$: $32$

\paragraph{Model}\mbox{}\\
Visitation prediction interval timesteps: $T_d = 6$\\
$\stopaction$ action threshold: $\kappa = 0.07$

\paragraph{General Learning}\mbox{}\\
Auxiliary objective weights: $\lambda_{\rm percept} = 1.0$, $\lambda_{\rm ground} = 1.0$, $\lambda_{\rm lang} = 0.25$ 

\paragraph{Supervised Learning}\mbox{}\\
Learning library: PyTorch 0.3.0\\
Optimizer: ADAM \\
Learning Rate: $0.001$\\
Weight Decay: $10^{-6}$\\
Batch Size: $1$

\paragraph{Imitation Learning}\mbox{}\\
Mixture decay: $\beta = 0.92$\\
Number of iterations: $100$\\
Number of environments for policy execution per iteration: $10$\\
Number of policy executions per iteration (executions): $47$ on average\\
Memory size (number of executions): $600$

\section{Proof of Theorem 5.1}

\begin{proof} 
Given that the state-visitation distribution of a policy $\pi: \stateset \rightarrow Pr(\actionset)$ is defined as $d(s; \pi, \mu) = \frac{1}{H}\sum_t d_t(\state; \policy, \mu)$, we can write the state-value function for the policy $\pi$ as:

\begin{small}
\begin{equation*}
V^\pi(s) = H \int d(s'; \pi, \delta_s) R(s') ds'\;\;,
\end{equation*}
\end{small}

where $\delta_s$ is the start-state distribution that places the entire probability mass on state $s$.

Using the definition $V^\pi(s)$ and assuming $\pi \in \Pi(\eta)$ we can write, 
\begin{eqnarray*}
 V^*(s) - V^\pi(s) &=& H \int d(s'; \pi^*, \delta_s) R(s')ds' - H \int d(s'; \pi, \delta_s) R(s')ds'\\
&=& H\int \left\{ d(s'; \pi^*, \delta_s) - d(s'; \pi, \delta_s)\right\} R(s')ds'\\
&\le& H\int \left\{ d(s'; \pi^*, \delta_s) - d(s'; \pi, \delta_s)\right\} (\tilde{R}(\phi(s')) + \alpha)ds'\\
&=& H\int \left\{ d(s'; \pi^*, \delta_s) - d(s'; \pi, \delta_s)\right\} \tilde{R}(\phi(s'))ds' + \\
&&H\alpha \int \left\{ d(s'; \pi^*, \delta_s) - d(s'; \pi, \delta_s)\right\}ds'\\
&=& H\int \left\{ d(s'; \pi^*, \delta_s) - d(s'; \pi, \delta_s)\right\} R(\phi(s'))ds'\\
&& \textrm{Because $d$ is a probability distribution, which gives } \\
&&\hspace{2em} \int \{ d(s'; \pi^*, \delta_s) - d(s'; \pi, \delta_s) \}ds'=\\
&&\hspace{6em}  \int d(s'; \pi^*, \delta_s)ds' - \int d(s'; \pi, \delta_s)ds'= 0\;\;.\\
&=& H \sum_{\tilde{\state} \in \tilde{\stateset}} \tilde{R}(\tilde{s})\int \mathbbm{1}\{\phi(s')=\tilde{s}\} \left\{ d(s'; \pi^*, \delta_s) - d(s'; \pi, \delta_s)\right\}ds' \\
&=& H\sum_{\tilde{s} \in \tilde{\stateset}} \tilde{R}(\tilde{s})\left\{ \tilde{d}(\tilde{s}; \pi^*, \delta_s) - \tilde{d}(\tilde{s}; \pi, \delta_s)\right\} 
\end{eqnarray*}

\begin{eqnarray*}
\hspace{92pt}&\le& H\sum_{\tilde{s} \in \tilde{\stateset}} \left|\tilde{R}(\tilde{s})\left\{ \tilde{d}(\tilde{s}; \pi^*, \delta_s) - \tilde{d}(\tilde{s}; \pi, \delta_s)\right\}\right|\\
&\le & H\left\{\sum_{\tilde{s} \in \tilde{\stateset}}  \left|\tilde{d}(\tilde{s}; \pi^*, \delta_s) - \tilde{d}(\tilde{s}; \pi, \delta_s)\right|\right\} \max_{\tilde{s} \in \stateset} |\tilde{R}(\tilde{s})| \\
&& \textrm{Using Holder's inequality}\\
&=& H\left\{\sum_{\tilde{s} \in \tilde{\stateset}}  \left|\tilde{d}(\tilde{s}; \pi^*, \delta_s) - \hat{d}(\tilde{s}) + \hat{d}(\tilde{s}) - \tilde{d}(\tilde{s}; \pi, \delta_s)\right|\right\} \max_{\tilde{s} \in \stateset} |R(\tilde{s})| \\
&\le & H\left\{\sum_{\tilde{s} \in \tilde{\stateset}}  \left|\tilde{d}(\tilde{s}; \pi^*, \delta_s) - \hat{d}(\tilde{s})\right| + \sum_{\tilde{s} \in \tilde{\stateset}} \left| \hat{d}(\tilde{s}) - \tilde{d}(\tilde{s}; \pi, \delta_s)\right|\right\} \max_{\tilde{s} \in \stateset} |R(\tilde{s})| \\
&\le & H\left(\sqrt{2 D_{KL}(\tilde{d}(\cdot; \pi^*, \delta_s) \mid \mid \hat{d})} + \sqrt{2D_{KL}(\hat{d} \mid \mid \tilde{d}(\cdot; \pi, \delta_s) )}\right) \max_{\tilde{s} \in \tilde{\stateset}} |\tilde{R}(\tilde{s})| \\
&& \textrm{Using Pinsker's inequality.}\\ 
&\le & H\left(\sqrt{2\epsilon} + \sqrt{2\eta} \right) \max_{\tilde{s} \in \tilde{\stateset}} |\tilde{R}(\tilde{s})| \\
&& \textrm{Using the theorem assumptions.}\\ 
&\le & H(R_{max} + \alpha)\left(\sqrt{2\epsilon} + \sqrt{2\eta} \right)\\
&& \textrm{Without loss of generality we assume $\tilde{R}(\tilde{s})=0$ for $\tilde{s}$ s.t. there exists no $s$},  \\
&& \textrm{where $\phi(s)=\tilde{s}$.  Additionally, $\tilde{R}: \tilde{\stateset} \rightarrow \mathbbm{R}^+$ rewards are only positive. } \\
&& \textrm{Therefore, } \tilde{R}(\phi(s)) \le R(s) + \alpha \Rightarrow \max_{\tilde{s} \in \tilde{\stateset}}|\tilde{R}(\tilde{s})| =  \max_{\tilde{s} \in \tilde{\stateset}}\tilde{R}(\tilde{s}) \le R_{max} + \alpha\;\;.
\end{eqnarray*}

We did not use any information about $\pi$ or $s$ in the above steps except for $\pi \in \Pi(\eta)$. Therefore taking supremum over $s$ and $\pi \in \Pi(\eta)$ completes the proof.
\end{proof}

\section{Additional instruction-following examples}

\begin{figure*}[t]
\centering
\includegraphics[width=\linewidth]{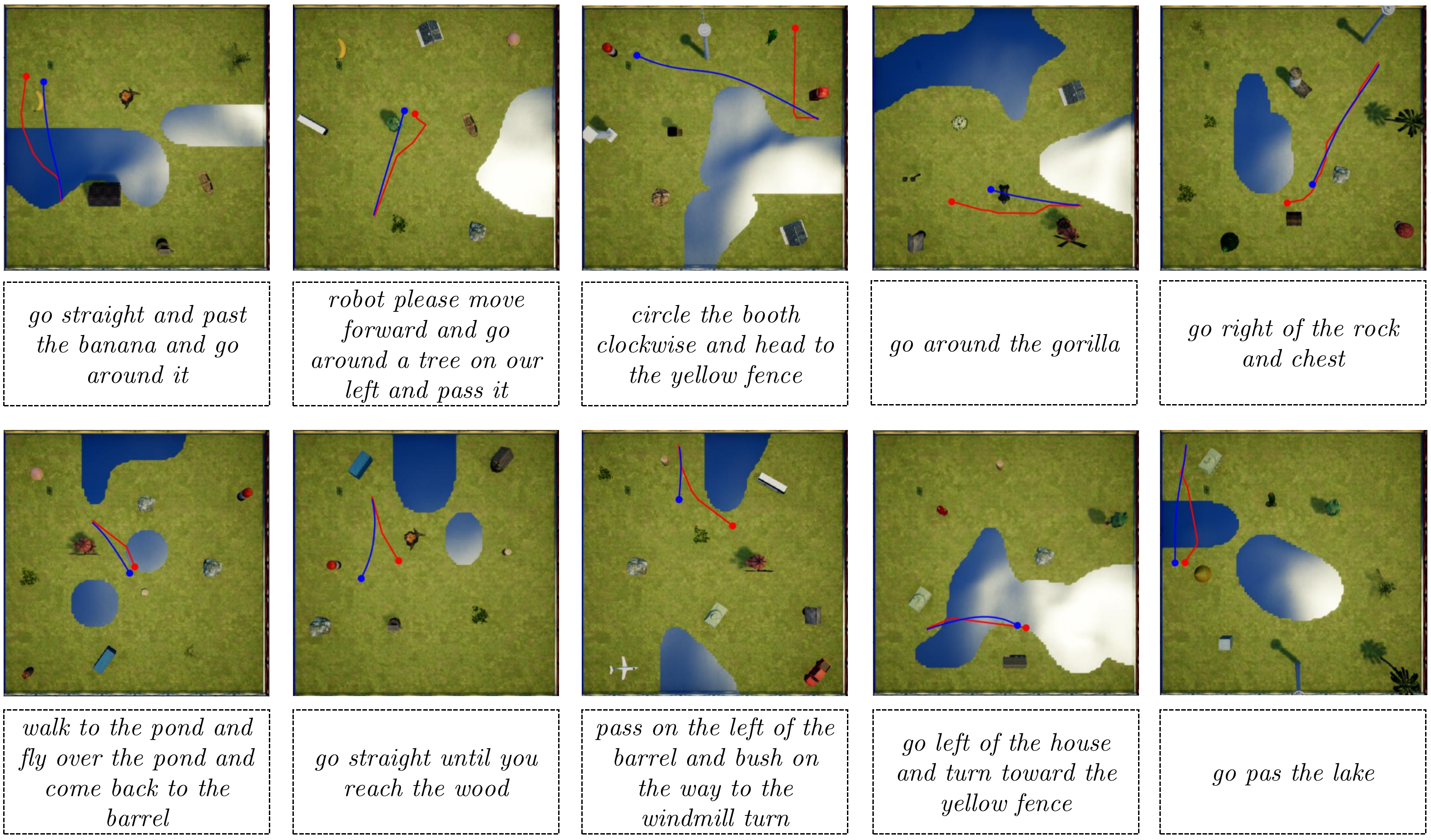}
\caption{Instruction following results (blue) and human demonstration trajectories (red)  on randomly selected instructions from the $\navdrone$ development set.}
\label{fig:extra_dev}
\end{figure*}

Figure~\ref{fig:extra_dev} shows example instructions from the development set along with the trajectories taken by our model and the human demonstrators.

\end{document}